%% file: PositNN.tex
\documentclass[conference]{IEEEtran}
\IEEEoverridecommandlockouts
\renewcommand\baselinestretch{0.99}
\usepackage{cite}
\usepackage{tabularx,ragged2e}
\usepackage{booktabs}
\newcolumntype{L}{>{\centering\arraybackslash}X}

\usepackage{enumitem}

\usepackage{multicol}
\usepackage{multirow}

\usepackage{subcaption}
\usepackage{graphicx}

\usepackage{algorithm}

\usepackage{algpseudocode}

\usepackage{url}

\algnewcommand\algorithmicforeach{\textbf{for each}}
\algdef{S}[FOR]{ForEach}[1]{\algorithmicforeach\ #1\ \algorithmicdo}
\algdef{SE}[DOWHILE]{Do}{doWhile}{\algorithmicdo}[1]{\algorithmicwhile\ #1}%
\algnewcommand{\LineComment}[1]{\Statex \(\///\) #1}

\usepackage{boldline}
\usepackage{amsmath}
\usepackage{amsfonts}
\usepackage{amssymb}
\usepackage{mathtools}

\usepackage{amsthm}

\usepackage{fixmath}
\usepackage{amsmath,amssymb,amsfonts}

\usepackage{graphicx}
\usepackage{textcomp}
\usepackage{xcolor}
\usepackage{comment}
\def\BibTeX{{\rm B\kern-.05em{\sc i\kern-.025em b}\kern-.08em
    T\kern-.1667em\lower.7ex\hbox{E}\kern-.125emX}}
\DeclareUnicodeCharacter{2212}{-}

\begin{document}
\bstctlcite{IEEEexample:BSTcontrol}
\title{\Large{Low-Precision Mixed-Computation Models for Inference on Edge} \\ 
\vspace{-0.5em}
}
\setlist[itemize]{noitemsep, topsep=0pt}
 \author{\IEEEauthorblockN{Seyedarmin Azizi, Mahdi Nazemi, Mehdi Kamal, and Massoud Pedram}
 \IEEEauthorblockA{Department of Electrical \& Computer Engineering, University of Southern California, Los Angeles, CA, USA\\
     \url{{seyedarm,mnazemi,mehdi.kamal,pedram}@usc.edu}}
 }

\maketitle

\vspace{-0.5em}
\input{abstract}

\section{Introduction}

\input{Introduction}

\section{Background}
\subsection{Posit Number System}
\input{Posit_Number_System}

\vspace{-0.5em}
\subsection{Quantization}
\input{Quantization}
\vspace{-0.3em}
\subsection{Posit-based Neural Networks}
\input{Posit-based_Neural_Networks}

\section{Proposed Mixed-Computation Framework}

\noindent
In the context of a neural network originally represented in floating-point format, our approach comprises two essential steps: the quantization of model parameters and activations, followed by a retraining process to restore accuracy. To achieve this, we employ a specialized algorithm that conducts layer-specific evaluations to determine whether to apply Posit number or fixed-point quantization selectively. Note that, for the sake of simplicity and without loss of generality, we employ Posit4 with $es=1$ in this study. 

Section A provides insights into the various Posit number variants employed, while Section B delves into the detailed layer-specific quantization process. In Section C, we introduce an innovative backward approximation technique tailored for Posit numbers, and Section D outlines the hardware architecture of our mixed-computation framework.

\vspace{-0.5em}
\subsection{Scaled Posit}
\input{ScaledPosit}

\subsection{Mixed-Computation Exploration}
\input{Mixed-Computation_Framework}

\subsection{Backward Gradient Approximation}
\input{Posits_Back_Propagation}

\subsection{Hardware Implementation}
\input{Hardware_Implementation}

\section{Results and Discussions}
\input{Results_and_Discussions}

\section{Conclusion}
This paper introduced a novel mixed-computation neural network processing approach designed for edge applications. By strategically utilizing both low-precision Posit and fixed-point number systems, the paper addresses challenges related to precision and energy consumption. Specifically, we employed 4-bit Posit for weights of high sensitivity and 4-bit FixP for other weights, guided by a heuristic method. Moreover, an efficient hardware implementation for MAC operations involving both Posit and FixP was also detailed. The paper's results confirm that this mixed-computation approach yields an average accuracy improvement of 1.5\% over fixed-point alone, albeit with a slight 0.19\% increase in energy overhead. This suggests that the trade-off between energy consumption and accuracy is reasonable, making the approach a viable alternative for resource-constrained edge applications.




\bibliographystyle{IEEEtran}
\bibliography{IEEEabrv,PositNN}
\end{document}

%% file: abstract.tex
\begin{abstract}
This paper presents a mixed-computation neural network processing approach for edge applications that incorporates low-precision (low-width) Posit and low-precision fixed point (FixP) number systems. This mixed-computation approach employs 4-bit Posit (Posit4), which has higher precision around zero, for representing weights with high sensitivity, while it uses 4-bit FixP (FixP4) for representing other weights. A heuristic for analyzing the importance and the quantization error of the weights is presented to assign the proper number system to different weights. Additionally, a gradient approximation for Posit representation is introduced to improve the quality of weight updates in the backpropagation process. Due to the high energy consumption of the fully Posit-based computations, neural network operations are carried out in FixP or Posit/FixP. An efficient hardware implementation of a MAC operation with a first Posit operand and FixP for a second operand and accumulator is presented. The efficacy of the proposed low-precision mixed-computation approach is extensively assessed on vision and language models. The results show that, on average, the accuracy of the mixed-computation is about 1.5\% higher than that of FixP with a cost of 0.19\% energy overhead.
\end{abstract}

\begin{IEEEkeywords}
Mixed-computation, Low-precision, DNN model, Posit, Fixed-point, Energy
\end{IEEEkeywords}

%% file: Introduction.tex
\noindent
Machine learning (ML) models, particularly deep neural networks (DNNs), have found extensive applications across various domains, including computer vision \cite{DBLP:conf/nips/KrizhevskySH12,  DBLP:conf/cvpr/HeZRS16}, and natural language processing \cite{radford2019language, DBLP:conf/naacl/DevlinCLT19}. With the increasing demand for preserving data privacy and addressing connectivity and latency concerns, ML on edge has gained significant attention. However, the rapid growth in the size and complexity of AI models poses substantial challenges for performing inference tasks on resource-constrained edge devices, necessitating advancements in the current state-of-the-art model compression techniques. Quantization, which allows for low bit-width representation \cite{noune20228, DBLP:conf/aaai/YuLSH021}, has been employed as the essential step for efficient and scalable deployment of neural networks.

Despite the significant improvements in quantization techniques, there is still non-negligible performance degradation at low bit-widths (e.g., 4 bits and lower) \cite{wang2018training,DBLP:journals/corr/abs-1805-06085,DBLP:conf/aaai/YuLSH021}, which hinders higher rates of compression, and thus limits the inference speed on edge devices. This can be justified by the high quantization error both in 4-bit floating point (FP4) and 4-bit fixed-point (FixP4) regimes. To tackle this issue, the Posit number system \cite{DBLP:journals/superfri/GustafsonY17} was introduced and proven to be superior to the floating-point number system because the Posit's dynamic range is significantly larger than the floating point's range, allowing for the representation of both very large and very small numbers with a relatively uniform precision across the entire range. In addition, Posit numbers offer gradual underflow and overflow behavior, which means that as a number approaches the limits of representable values, the precision degrades gracefully. The combination of these two features leads to higher accuracy during inference \cite{DBLP:journals/access/NambiUSLMK21}. In fact, it is shown that 8-bit Posit numbers deliver the same performance as FP32 in neural networks \cite{lu2020evaluations}. However, adopting Posit numbers in DNNs has its own drawbacks. Since Posit uses a complex representation, its arithmetic units are slower and consume more energy. Although there have been prior efforts to improve the energy efficiency of Posit implementation\cite{DBLP:journals/ipsj/NakaharaMKAI22, DBLP:journals/tetc/MurilloBBKKB22}, Posit hardware is still more costly than fixed-point (FixP) or floating-point (FP) hardware.

To overcome the limitations inherent in existing numerical representations within neural networks, we have devised a novel hybrid computational approach that leverages the strengths of both Posit and FixP number systems. In our system, we strategically employ 4-bit Posit (Posit4) with an exponent field width of 1 ($es=1$) to represent the parameters (weights) of layers that are highly sensitive to quantization errors. This capitalizes on Posit's superior accuracy in handling such cases. For layers that are less susceptible to quantization errors, we have opted for the more hardware-efficient FixP4 representation.
To mitigate the computational cost and energy overhead associated with Posit, we have chosen to represent all layer activations in our proposed mixed-computation approach using the FixP4 number system. This combined strategy results in our mixed-computation system achieving superior model performance in 4-bit quantization scenarios while incurring minimal computational overhead.

In our approach to selecting the appropriate number system for each layer, we introduce a sensitivity analysis algorithm. This algorithm takes into account the average magnitude of the gradients of the output loss function concerning layer weights and quantization errors. Additionally, we outline the training mechanisms for our newly developed mixed-computation models. This innovation enables the quantization of well-established vision and language models into 4-bit Posit numbers for the very first time.
Additionally, we present a specialized hardware implementation that employs a Posit decoder-only design, thereby eliminating the resource-intensive encoder component. This modification further simplifies the integration of Posit arithmetic into existing hardware architectures.
Our comprehensive analysis sheds light on the trade-offs associated with choosing between Posit and fixed-point quantization methods for accelerating DNN inference. 
\noindent
Based on above explanations, the contributions of this work can be summarized as below:
\begin{itemize}[noitemsep,topsep=0pt]
    \item Introducing a framework for developing low-precision mixed-computation models for edge applications. 
    
    \item Presenting a specialized training methodology that leverages a customized backpropagation algorithm for the parameters' update in layers utilizing Posit numbers. 
    \item Proposing an efficient 4-bit Posit/FixP multiply-accumulator (MAC) hardware structure. 
    \item Evaluating the proposed approach under various vision and language models.
\end{itemize}


%% file: Posit_Number_System.tex
\begin{figure}[htb]
    \centering
    \includegraphics[width=0.8\columnwidth]{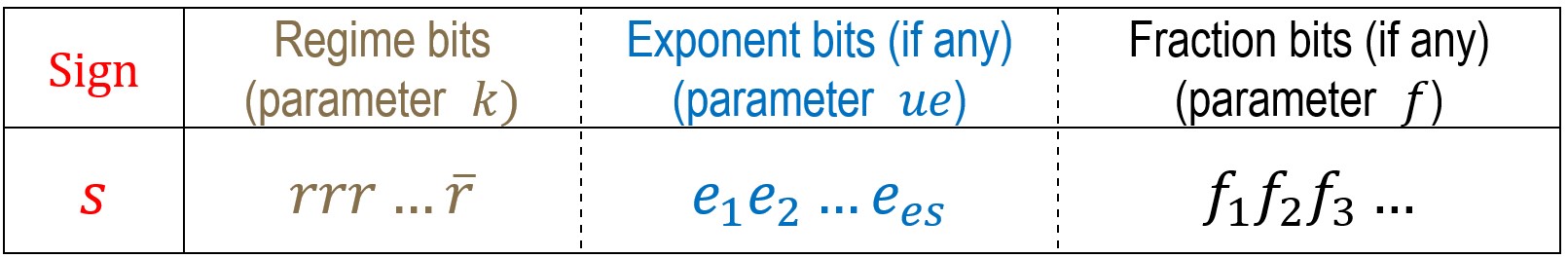}
    \caption[Parallel tree-based reduction]{General structure of a number in the Posit NS and its corresponding value -- read $1.f$ as $(1+f)$.}
    \label{fig:PositGeneral}
    \vspace{-0.5em}
\end{figure}

\noindent
The Posit numerical format has emerged as a compelling alternative to traditional floating-point arithmetic \cite{DBLP:journals/superfri/GustafsonY17}. 
In this system, an n-bit number is denoted as $P(n,es)$, where $es$ stands as a control parameter that defines the bit width of the exponent section. A distinguishing feature of Posit numbers in contrast to FP numbers is the incorporation of tapered precision, which paves the way for both gradual underflow and overflow. Intriguingly, Posits avoid the extremes of overflowing to infinity or underflowing to zero, as highlighted in \cite{DBLP:journals/superfri/GustafsonY17}. Subsequent studies \cite{30711,8766229} further detail the merits of Posits, underscoring enhanced dynamic range, accuracy, and cross-machine consistency compared to the conventional FP systems.
The format for representing Posit numbers is depicted in Figure \ref{fig:PositGeneral}. The formula used to calculate the numerical value of a Posit number $p$ is as follows: $val(p) = \{ 0 \textnormal{ if } p = 00 \cdots 0 \textnormal{; } \pm \infty \textnormal{ if } p = 10 \cdots 0 \textnormal{; } (-1)^s \times ({2^2}^{es})^k \times 2^{ue} \times 1.f \textnormal{ otherwise} \}$. 

The parameter \(k\) is equal to \(−m\) or \(m-1\), where \(m\) denotes the number of running 0's or 1's in the regime field, respectively.
Posit allows control over dynamic range and precision by adjusting the exponent field width ($es$) \cite{7-gustafson2017posit}.

Despite the advantages conferred by the Posit numerical format in comparison to FP and FixP number systems, hardware implementation of Posit arithmetic remains a non-trivial task. This complexity arises primarily from the quadripartite structure of Posit representation, of which three fields possess dynamically adjustable, non-constant bit widths to accommodate a wide dynamic range. Arithmetic operations involving Posit numbers typically necessitate the employment of a specialized fixed format, referred to as the "raw format." Conversion between the raw and Posit formats requires decoding and encoding processes, contributing to computational overhead. As reported in \cite{DBLP:conf/socc/LuLWFLWD19}, the energy overhead of encoding and decoding in MAC operation could be around 40\%.

%% file: Quantization.tex

Truncating the operand bitwidth presents an effective strategy for achieving considerable energy savings, particularly during the data-loading and data-storing phases. To leverage these energy savings, researchers have proposed the use of low-precision number formats. Examples include 8-bit floating-point (FP8) \cite{wang2018training, noune20228, micikevicius2022fp8} and 8- or even 4-bit fixed-point (FixP8 or FixP4) or integer numbers \cite{dong2019hawq}.
As an alternative approach, non-uniform quantization could lead to better performance and smaller model sizes, but it suffers from inefficient hardware realization and may need specialized hardware optimizations. 

%% file: Posit-based_Neural_Networks.tex
\noindent
Generally, the parameters of DNNs tend to exhibit a normal distribution with a mean value of zero, largely owing to regularization techniques \cite{noune20228, banner2018scalable}. Consequently, using FixP representation, which maintains a uniform distribution within its dynamic range, for low-precision training poses significant challenges \cite{DBLP:journals/corr/abs-2206-02915}. Given this non-uniform characteristic, Posit emerges as a promising number system for the design of DNN accelerators \cite{carmichael2019performance}.
Posit has found applications in various works aimed at accelerating neural networks. For instance, \cite{DBLP:journals/corr/abs-1908-02386} presented a software-hardware co-design approach for low-precision DNN training and inference while optimizing for the Energy-Delay product of FPGA hardware. In \cite{carmichael2019deep, carmichael2019performance}, a set of customized Exact MAC units designed for Posit computations were introduced and evaluated for the inference phase of small-sized neural networks.
In addition to inference, several studies including \cite{lu2020evaluations, murillo2020deep, DBLP:journals/corr/abs-1908-02386, DBLP:conf/socc/LuLWFLWD19}, have explored the impact of using Posit in the training phase of neural networks. For instance, Deep PeNSieve \cite{murillo2020deep} employed Posit(32,2) and Posit(16,1) for training and gradient updates, while using Posit(8,0) for the inference stage.
However, prior research on Posit-based DNNs has struggled to achieve promising results below the Posit8 format, especially on well-known vision models or large datasets, due to significant quantization errors.
%


%% file: ScaledPosit.tex
\noindent
Quantizing with Posit can be seen as a mapping function from floating-point numbers to the Posit values. More concretely, in Posit4 system, this would be an encoding from floating-point numbers to the nearest value in the vector defined using Posit4:
\begin{equation}
\begin{aligned}
    Posit4 = &[-16, -4, -2, -1, -0.5, -0.25, - 0.0625, \\
             &  0, 0.0625, 0.25, 0.5, 1, 2, 4, 16]
\end{aligned}
\end{equation}
\begin{equation}
    Q^{Posit}(w) =  \underset{x \in Posit4}{\arg\min}|w -x|
\end{equation}
Certain elements in the Posit4 vector deviate substantially from the typical range of weight values encountered in DNNs (e.g., 16). In DNNs, the weights commonly exhibit a mean value around zero and a standard deviation close to one \cite{DBLP:conf/cvpr/XieXP17}. Such a discrepancy can result in a suboptimal utilization of the Posit numerical format. To ameliorate this issue, we propose to employ scaled values of the Posit4, divided by 4 (\(Posit4_{sc,4}\)) and 8 (\(Posit4_{sc,8}\)). In this case, the range of Posit values will be close to the distribution of the weights in DNNs. The integration of these scaling factors into the Posit decoding process is simple, and since the considered scaling factors are the power of 2 numbers, only the decimal point in the raw posit format is moved to the left.

%% file: Mixed-Computation_Framework.tex
\noindent
Since the parameters of the DNN layers may have different distributions (different dynamic range and precision around zero) \cite{dong2019hawq,DBLP:journals/corr/abs-2308-06422}, to exploit the mixed-computation paradigm fully, we must determine which neural network layers should be mapped to \(Posit_{sc,4}\) or \(Posit_{sc,8}\) values and which ones should be quantized using fixed-point representation. This section showcases how we can identify the layers that would benefit more from Posit representation. Intuitively, we should look for the layers in which mapping to Posit would reduce the quantization error more when compared to fixed-point. However, considering quantization error solely may be misleading since the effect of quantization error of different network layers on the output loss function is not the same \cite{dong2019hawq}. To take into account both the quantization error and the significance of the error's effect on the loss, we can use the first-order Taylor expansion of the neural network's loss. Assume we are only quantizing layer \(l\) of a model, and all other layers are kept unchanged. If we denote the layer parameters as \(\mathbold{w_l}\), we can expand the model's loss \(\mathcal{L}\) as a function of \(\mathbold{w_l}\):
\begin{equation}
    \label{eq1}
    \begin{split}
        \mathcal{L}(Q(\mathbold{w}_l)) \approx \mathcal{L}(\mathbold{w}_l) & + (Q(\mathbold{w}_l) - \mathbold{w}_l)^\mathrm{T} \mathbold{\nabla}\mathcal{L}_{\mathbold{w}_l}
    \end{split}
\end{equation}
where \(Q\) is the quantization function. As seen from this equation, if we think of quantization as a perturbation to the trained parameters of layer \(l\), the extent to which the output loss is affected as the result of this perturbation is equal to \(||(Q(\mathbold{w}_l) - \mathbold{w}_l)^\mathrm{T} \mathbold{\nabla}\mathcal{L}_{\mathbold{w}_l}||\). Consequently, to exploit Posit effectively, we should apply it to layers that (i) have higher gradient values, and (ii) the reduction in quantization error is significant (compared to fixed-point). Thus, we introduce a compound \textit{sensitivity metric} (\(s\)) that considers both factors. This metric, denoted by \(s_l\) for \textit{$l^{th}$} layer, is calculated as:
\begin{equation} 
    \label{eq1}
    \begin{aligned} 
    & s_{l_{sc,k}}  = \\
    & \frac{(||Q^{FixP}(\mathbold{w}_l) - \mathbold{w}_l|| - ||Q^{Posit}_{sc,k}(\mathbold{w}_l) - \mathbold{w}_l||) \times ||\mathbold{\nabla}\mathcal{L}_{\mathbold{w}_l}||}{n_l}
    \end{aligned}
\end{equation}
\begin{equation}
    \label{eq2}
    s_l = max(s_{l_{sc,8}}, s_{l_{sc,4}})
\end{equation}
where \(Q^{FixP} \) and \(Q^{Posit}_{sc,k}\) denote the fixed-point and Posit quantization functions with scale k, respectively, and \(n_l\) is the number of parameters in layer \(l\). In essence, the difference in quantization errors serves as an indicator of the extent to which Posit numbers outperform fixed-point numbers, while the magnitude of the gradient conveys the level of influence that a particular layer wields over the overall output.
The (\ref{eq2}) determines whether scaling by a factor of four or eight is suitable for layer \(l\).  We normalize \(s_l\) by \(n_l\) to obtain the average quantization error for each layer.

Typically, we prefer to employ Posit for the layers with high sensitivity \(s\) values since the higher the value of \(s_l\), the more significant the impact of using Posit in layer \(l\); thus, we compute this quantity for all layers of the model and sort the layers based on it. Subsequently, we start applying Posit to the layers based on the sorted list until the total number of parameters encoded to Posit numbers exceeds a predefined percentage \(\eta\) of the total number of parameters in the given neural network. This limit is set based on the tolerable computational overhead of using Posit numbers. 

\begin{figure}[hbt]
\vspace{-1em}
\centering
\begin{subfigure}{0.23\textwidth}
    \includegraphics[width=\textwidth]{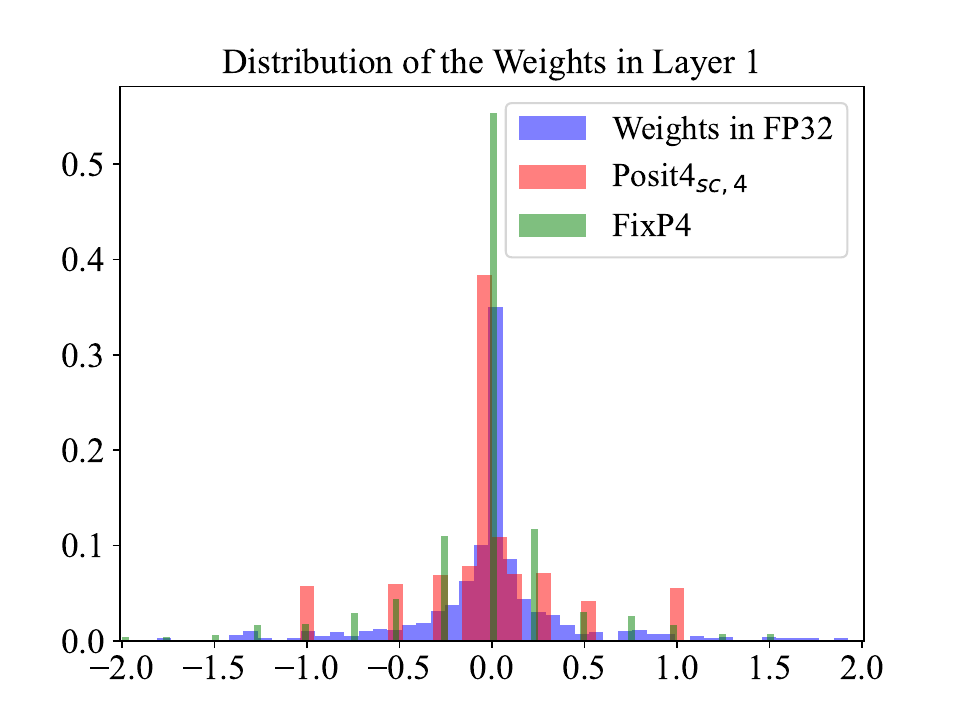}
    \caption{ \scriptsize Quantization Error associated with \(Posit4_{sc,4}\), FixP4, and \(Posit4_{sc,8}\) are 0.019, 0.028, and 0.043, respectively.}
    \label{fig:first}
\end{subfigure}
\hfill
\begin{subfigure}{0.23\textwidth}
    \includegraphics[width=\textwidth]{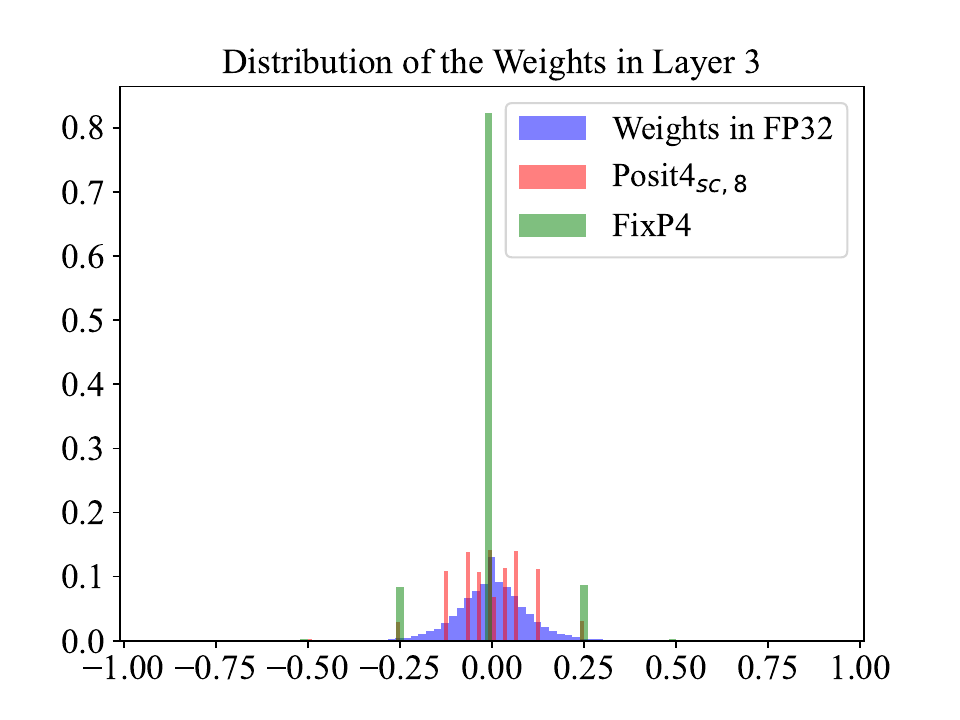}
    \caption{ \scriptsize Quantization Error associated with \(Posit4_{sc,8}\), FixP4, and \(Posit4_{sc,4}\) are 0.0004, 0.0042, and 0.0027, respectively.}
    \label{fig:second}
\end{subfigure}
\hfill       
\caption{Distributions of the parameters of two layers in MobileNetV2 under different number systems.}
\label{fig:distributions}
\vspace{-1em}
\end{figure}

\begin{algorithm}[tb]
\footnotesize
  \caption{Mixed-Computation Model}
  \label{alg:mixed-computation}
  \begin{algorithmic}[1]
    \Require
        \Statex  \(\mathbold{\Theta}\) \Comment{Full-precision trained DNN to be quantized}
        \Statex  \(\mathcal{D}\) \Comment{Dataset}
        \Statex \(\eta = 0.1\)
            
    \Ensure
        \Statex \(\mathbold{\Theta}^q\) \Comment{Quantized neural network}

    \Statex
    
    \State \(\texttt{gradients} = \texttt{do\_back\_propagation}(\mathbold{\Theta},\mathcal{D})\)

    \State \(\texttt{Layers} = \mathbold{\Theta}.layers()\)
    \State \(N = \texttt{number\_of\_parameters}(\texttt{Layers})\)

    \State \(\texttt{PositLayers} = []\) \Comment{Layers that should be mapped to Posit}
    \State \(\texttt{S} = []\) \Comment{List of \(s\) values}
    \ForEach {\(\texttt{layer}\)  \(\in\) \(\texttt{Layers}\)} 
        \State \({\texttt{S.add}(\texttt{compute\_{s\_l}}(\mathbold{\Theta}, \texttt{layer}, \texttt{gradients}))}\)
    \EndFor
    \State \(\texttt{S} = \texttt{Sort(S)}\)
    \State \(\texttt{Layers} = \texttt{Sort(Layers(S.indexes()))}\)
    \Do
        \State \( \texttt{PositLayers.add(Layers.pop())}\)

    \doWhile {\(\texttt{number\_of\_parameters}(\texttt{PositLayers}) \leq \eta * N\)}
    
    \State \( \mathbold{\Theta}^q = \texttt{Quantize}(\mathbold{\Theta}, \texttt{Layers}, \texttt{PositLayers})\)
    \State \( \mathbold{\Theta}^q = \texttt{Train}(\mathbold{\Theta}^q, \mathcal{D})\)
    \State{\textbf{return} \(\mathbold{\Theta}^q \)}

\end{algorithmic}
\end{algorithm}
Algorithm \ref{alg:mixed-computation} summarizes our method for mapping a trained model to a mixed-computation one. The quantization function maps the layers in \textit{PositLayers} to either \(Posit4_{sc,4}\) or the \(Posit4_{sc,8}\) based on (\ref{eq2}), and the rest of the layers are mapped to FixP4.
For illustration, refer to Figure \ref{fig:distributions}, which showcases the weight value distributions in the first and third layers of the trained MobileNetV2 architecture. These specific layers have been designated for mapping to Posit representations using Algorithm \ref{alg:mixed-computation}. An examination of the plot reveals that the original weight distribution closely approximates a Gaussian distribution centered around zero. It is important to emphasize that the quantization error for Posit representation is expected to be lower than that for FixP4. This can be attributed to the higher concentration of Posit values near zero, enabling a more precise representation of weight values. In contrast, FixP4 exhibits a uniform distribution around zero, resulting in a less optimal capture of the inherent characteristics of the weight distribution.
Moreover, since the distribution of the weights in layer 3 is denser than that of layer 2, \(Posit4_{sc,8}\) leads to a smaller quantization error for layer 3, and \(Posit4_{sc,4}\) better fits the layer 1, and that's why they have been selected in (\ref{eq2}).

The quantization of weight parameters represents a key component of our approach, demanding a careful examination. When it comes to weight quantization in the FixP4 format, we adopt an entropy-based uniform quantization algorithm outlined in \cite{DBLP:conf/cvpr/LiuCHXS22}. This particular method has demonstrated its robustness across a range of computational scenarios. However, in contrast to the approach proposed by \cite{DBLP:conf/cvpr/LiuCHXS22}, where weight values are confined within the \(-1\) to \(+1\) range, we introduce a modification to allow for a flexible combination of fractional and integral bit-widths. Specifically, we determine the lower and upper saturating thresholds based on the selected bit widths. The mathematical representation of this quantization process is as follows:
\begin{equation} \small
\begin{aligned}
{
  scale = mean(abs(W)) } \times \frac{2^n -1}{2^{n-1}}
\end{aligned}    
\end{equation}
\begin{equation} \small
\begin{aligned}
   {\hat{W} = round((clip(\frac{W}{scale}, low, high) - low) \times \frac{2^n -1}{high-low})
}
\end{aligned} 
\end{equation}
\begin{equation} \small
\begin{aligned}
   { Q^{FixP}(W) = \hat{W} * \frac{high-low}{2^n -1} + low
}
\end{aligned} 
\end{equation}
where $Q^{FixP}()$ is the fixed-point quantization function and $W$ is the input weight. 
Through empirical evaluation, we have determined that a bit allocation of two bits for the integral part and two bits for the fractional part consistently delivers optimal performance when employing the FixP4 architecture across a wide range of models and datasets. In this configuration, the scale factor for quantization is computed as the mean absolute value of the weight tensor \(W\). By dividing the tensor by this scale factor, some of the larger weight values may approach the saturating thresholds. However, note that smaller weights, which constitute a significant portion of the weight distribution, experience a relative increase in magnitude. This adjustment promotes a more uniform distribution of smaller weights across the quantization bins. Empirical evidence supports that this mechanism maximizes the retention of crucial information during the fixed-point quantization process, as discussed in \cite{DBLP:conf/cvpr/LiuCHXS22}.

In the context of quantizing activations, we have adopted the Parameterized Clipping activation (PACT) approach as introduced in \cite{DBLP:journals/corr/abs-1805-06085}. This approach allows us to recover accuracy after quantization by training a clipping threshold, which is expressed as follows:
\begin{equation}
y = PACT(x) = 0.5 \left(|x| - |x-\alpha| + \alpha\right)
\end{equation}
After obtaining the clipped values using this method, we proceed to quantize them using the fixed-point format:
\begin{equation}
x^q = \text{round}\left(y \times \frac{2^n-1}{\alpha}\right) \times \frac{\alpha}{2^n-1}
\end{equation}
Retaining activations across all layers in the FixP4 representation allows us to perform computations in the fixed-point format, even for layers whose parameters are originally mapped to Posit4. 

%

%% file: Posits_Back_Propagation.tex
\noindent

\begin{figure}
\centering
\includegraphics[scale=0.80]{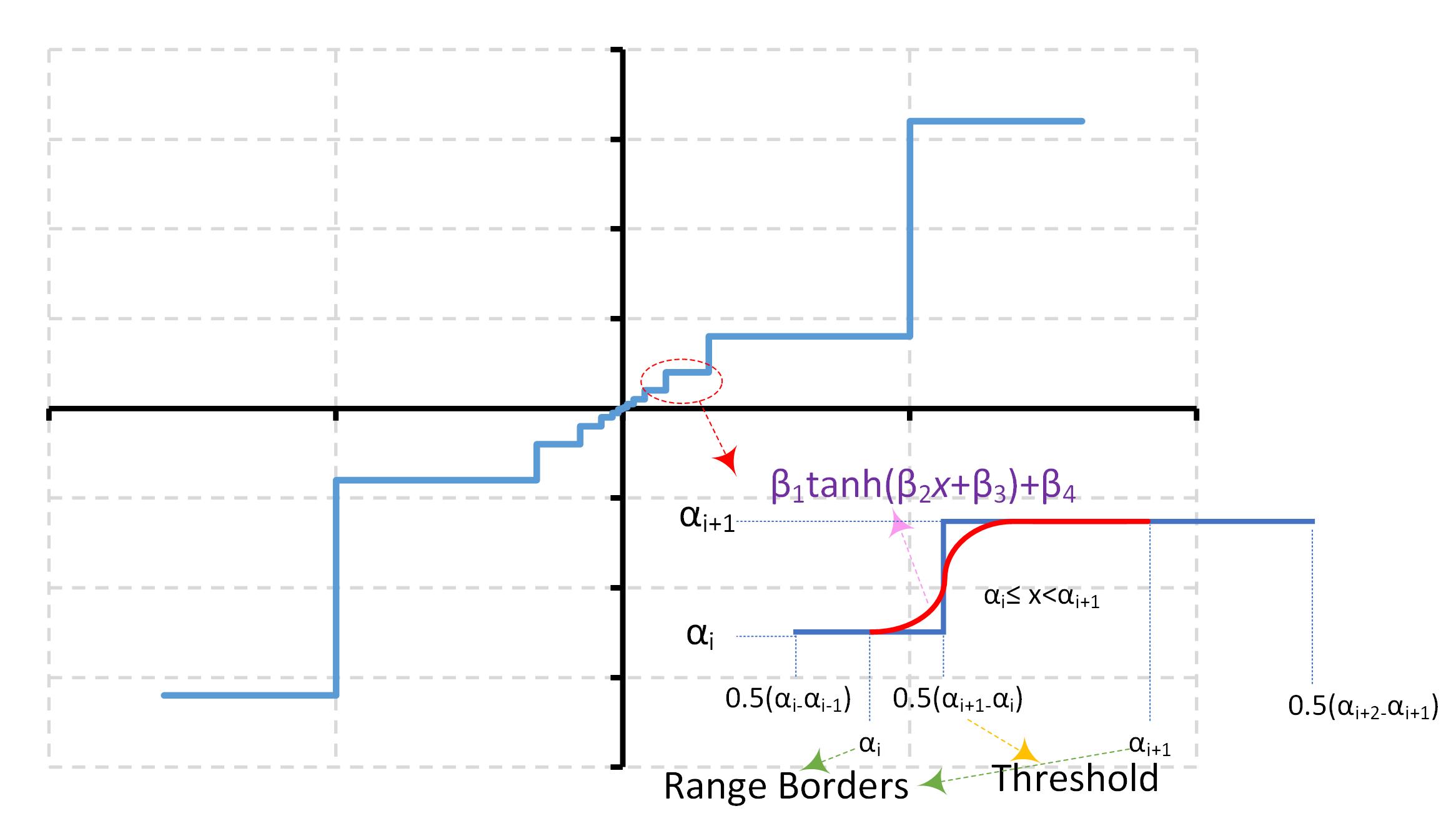}
\caption{Proposed gradient estimator of the Posit quantizer.} 
\vspace{-2em}
\label{prop:STE}
\end{figure}

When dealing with weights represented using the uniform quantization (e.g., FixP format), we employ the standard Straight-Through Estimator (STE) method, as described in \cite{DBLP:journals/corr/BengioLC13} to compute the gradient of the quantization module with respect to its input.
However, for Posit weights, the quantization approach is non-uniform, and we introduce an alternative approximation method. In the Posit quantizer, the real input value \(x\) is rounded to the nearest representable value within the chosen Posit number system. Assuming a set of representable values ${\alpha_{-7}, \alpha_{-6}, ... \alpha_{0}, ..., \alpha_{6}, \alpha_{7}}$ in Posit4, real input values within the range $[\alpha_{i}, \alpha_{i+1}]$) close to $0.5(\alpha_{i+1} - \alpha_{i})$ (i.e., the threshold value) exhibit a larger difference from their corresponding quantized values (i.e., higher $|x-x_q|$) while this difference diminishes for input values near the Posit representable numbers (e.g., $\alpha_i$). In other words, the gradient for the input values (in the range of $[\alpha_{i}, \alpha_{i+1}]$) near the $0.5(\alpha_{i+1} - \alpha_{i})$ should be larger than others, and the gradients of the values close to the range boundaries should be nearly zero. Based on these observations, we propose to use the $tanh(x)$ function (see Figure \ref{prop:STE}) to approximate the backward gradients in the Posit quantizer ($\textit{Q}^{Posit}_G()$). For a full precision weight in the range of $[\alpha_{i}, \alpha_{i+1}]$, its approximate gradient is defined by:
\begin{equation} \small
\begin{aligned}
    {\partial{Q^{Posit}_G} \over \partial{x}}={{\partial} \over \partial{x}}( {2 \over \alpha_{i+1} - \alpha_{i}} tanh({5 \over {\alpha_{i+1} - \alpha_{i}}} (x - 0.5(\alpha_{i+1} - \alpha_{i}))))
\end{aligned}
\end{equation}
Note that in contrast to STE and G-STE \cite{DBLP:conf/cvpr/LiuCHXS22}, which account for similar gradients across the entire range, our approach primarily concentrates on values around the thresholds while largely disregarding the gradients of values near the range boundaries. 

%% file: Hardware_Implementation.tex
The hardware for performing Posit arithmetic operations tends to consume more energy and exhibit higher delays compared to their FixP counterparts. However, it's worth noting that the primary arithmetic operation required for DNN accelerators is the Multiply-Accumulate (MAC) operation, which plays a crucial role in executing dot-product operations.

\begin{figure}
\centering
\includegraphics[scale=0.8]{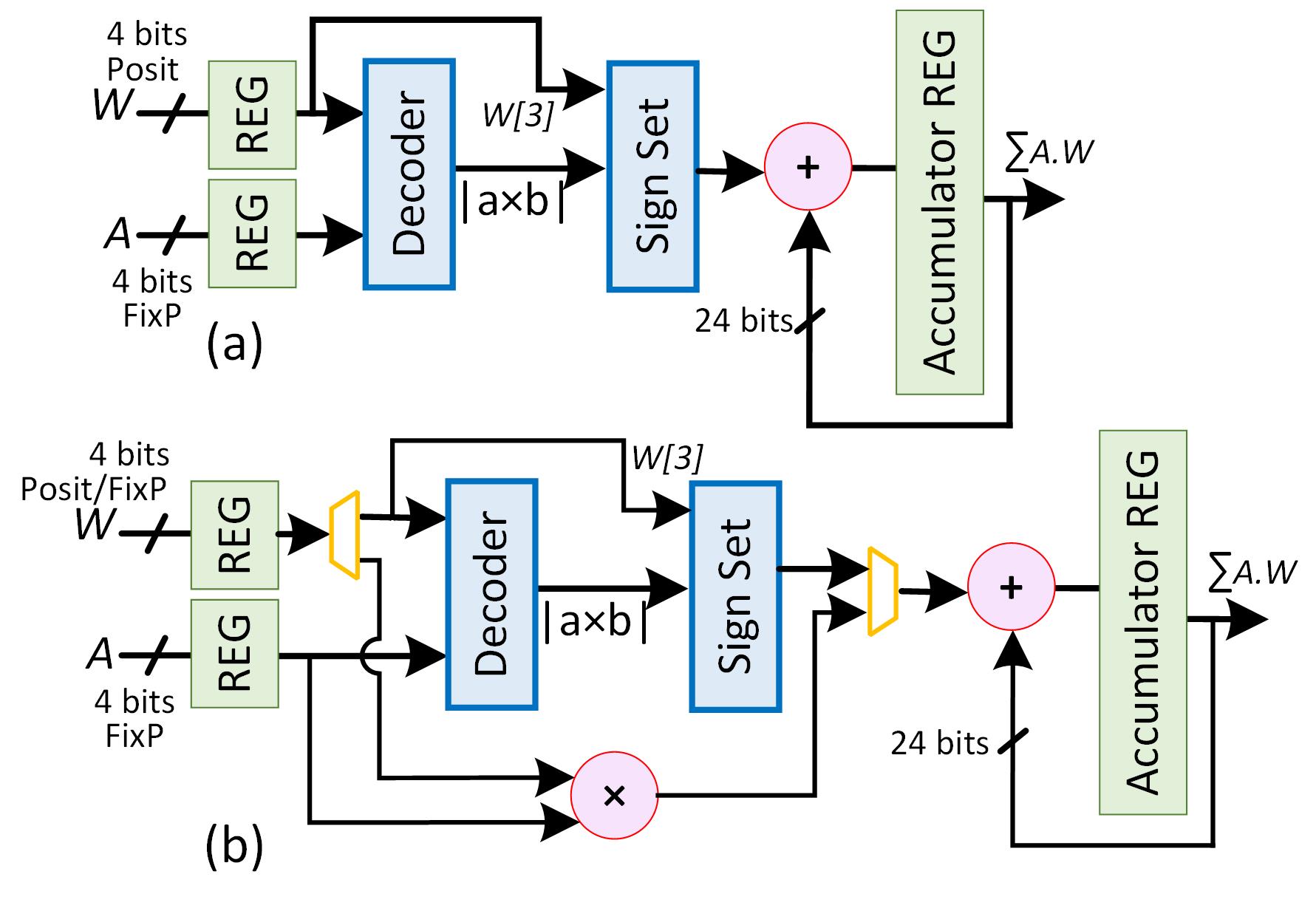}
\caption{Posit/FixP MAC unit.} 
\label{prop:MAC}
\vspace{-2em}
\end{figure}

In our proposed mixed-computation approach, where only the weights are in Posit format, we recommend a Posit/FixP MAC hardware structure (refer to Figure \ref{prop:MAC}(a)). This structure involves one of its input operands being in Posit4 format, while the other is in FixP4 format. Given that the Posit operand consists of only 4 bits, we utilize a lookup table (LUT) to obtain the raw format of the input Posit number. Moreover, as the raw format of positive Posit4 values contains only one '1' bit, the multiplication operation 
$W \times A$ can be simplified to a shift operation. Therefore, we incorporate the multiplication operation within the Decoder unit, with its output being $|W \times A|$. Subsequently, based on the value of the last bit of the Posit operand, we determine the sign of 
$W \times A$ using the Sign Set unit, which outputs the result in two's complement format. It's important to note that activations are treated as positive numbers. Finally, the multiplication result is accumulated with the intermediate result of the dot-product operation.

The raw format of Posit4 requires 10 bits, resulting in the output of the multiplication operation being 14 bits wide. Additionally, we allocate 10 extra bits for accumulation. Consequently, the width of the add operation is set to 24 bits. In comparison, for a fully FixP MAC operation with the same 10 extra bits allotted for accumulation, the width of the add operation is 18 bits.
In our mixed-computation approach, it is possible to have a subset of weights in Posit format while the remainder are in FixP format. Therefore, we combine a fully FixP MAC with the Posit/FixP MAC to be utilized in DNN accelerators. The proposed combined MAC structure is depicted in Figure \ref{prop:MAC}(b).

Based on our investigations, using 45nm technology, the energy consumption of the Posit/FixP MAC operation is approximately 30\% higher than that of the fully FixP MAC operation. Consequently, in the worst-case scenario where all weights are in the Posit number system, the MAC operation contributes a 30\% energy overhead to the total energy consumption of computations. However, it's worth noting that as reported in studies such as \cite{7551399} and \cite{8686088}, computation operations typically account for less than 10\% of the total energy consumption of DNN accelerators. Furthermore, in our proposed mixed-computation approach, some weights are represented using FixP rather than Posit. This results in the overall energy overhead from Posit computations being negligible.

%% file: Results_and_Discussions.tex
We evaluated the effectiveness of our proposed mixed-computation approach across various vision and language models, with a focus on tasks involving the CIFAR-10, CIFAR-100, Tiny ImageNet, and ImageNet datasets. To explore mixed-computation, we set the \(\eta\) threshold at 10\%. After loading the pre-trained 32-bit floating-point model checkpoints, we applied Algorithm \ref{alg:mixed-computation} to determine the quantization mode for each layer. Subsequently, we employed PyTorch for retraining the quantized models, conducting training runs lasting 90 epochs. We utilized the Adam optimizer with a batch size of 128 and implemented Cosine scheduling for learning rate adjustments.
Table \ref{tab:vision models} presents the Top-1 accuracy results for different computational paradigms across well-established vision models. Posit4 and Mixed-Computation consistently outperform FixP4, largely owing to their denser numerical representations in proximity to pre-trained weight values. Notably, Mixed-Computation achieves nearly optimal performance by applying Posit mapping to fewer than 10\% of the model parameters. For instance, in the case of models like ResNet-20 and ResNet-18 \cite{DBLP:conf/cvpr/HeZRS16}, Mixed-Computation approaches the performance of a fully Posit-based design with minimal computational overhead. Remarkably, in models such as MLP-Mixer \cite{tolstikhin2021mlp}, the performance gap between FixP4 and Mixed-Computation is substantial, amounting to 6.5\%.

For language tasks, we evaluated the efficacy of Mixed-Computation using BERT\cite{DBLP:conf/naacl/DevlinCLT19} and GPT-2 \cite{radford2019language} as foundational models. BERT was evaluated on the MNLI dataset for Natural Language Inference and the SST-2 dataset for sentiment classification. Likewise, GPT-2 was assessed using the WikiText-2 dataset for text generation tasks. Across all evaluations, Mixed-Computation significantly outperforms fixed-point quantization with negligible computational overhead. The benefits of employing low bit-width Posit representations are particularly visible in these memory-constrained models.

\begin{table}[tb]
\centering
\caption{{Mixed-Computation Performance on Vision Models }}
\label{tab:vision models}
\resizebox{\columnwidth}{!}{
\begin{tabular}{c c c c c c}
    \toprule
    \multirow{2}{*}{\textbf{Dataset}}   & \multirow{2}{*}{\textbf{Model}}    & \textbf{Full-Precision (FP32)} & \textbf{FixP4}         & \textbf{Posit4}         &  \textbf{Mixed-Computation} \\
    {}                                  & {}                                        & \textbf{(\%)}  & \textbf{(\%)}             & \textbf{(\%)}     & \textbf{(\%)} \\
    \midrule[\heavyrulewidth]
    \multirow{2}{*}{\textbf{CIFAR-10}}    & \multirow{2}{*}{\textbf{ResNet-20}}    &
    \multirow{2}{*}{91.5}    & \multirow{2}{*}{90.2}    & \multirow{2}{*}{91.3}    & \multirow{2}{*}{91.1} \\
    {} & {} & {}    & {}    & {}    & {} \\

    \midrule[\heavyrulewidth]
    \multirow{2}{*}{\textbf{CIFAR-10}}    & \multirow{2}{*}{\textbf{VGG-16}}    &
    \multirow{2}{*}{91.8}    & \multirow{2}{*}{91.4}    & \multirow{2}{*}{92.5}    & \multirow{2}{*}{91.8} \\
    {} & {} & {}    & {}    & {}    & {} \\
    \midrule[\heavyrulewidth]

    \multirow{2}{*}{\textbf{CIFAR-10}}    & \multirow{2}{*}{\textbf{MLP-Mixer}}    &
    \multirow{2}{*}{94.2}    & \multirow{2}{*}{74.7}    & \multirow{2}{*}{83.4}    & \multirow{2}{*}{81.2} \\
    {} & {} & {}    & {}    & {}    & {} \\
    \midrule[\heavyrulewidth]
    
    \multirow{2}{*}{\textbf{CIFAR-100}}    & \multirow{2}{*}{\textbf{ResNet-18}}    &
    \multirow{2}{*}{75.4}    & \multirow{2}{*}{73.6}    & \multirow{2}{*}{74.9}    & \multirow{2}{*}{74.4} \\
    {} & {} & {}    & {}    & {}    & {} \\
        \midrule[\heavyrulewidth]
    \multirow{2}{*}{\textbf{CIFAR-100}}    & \multirow{2}{*}{\textbf{MobileNetV1}}    &
    \multirow{2}{*}{65.5}    & \multirow{2}{*}{63.8}    & \multirow{2}{*}{64.5}    & \multirow{2}{*}{64.1} \\
    {} & {} & {}    & {}    & {}    & {} \\
        \midrule[\heavyrulewidth]
    \multirow{2}{*}{\textbf{CIFAR-100}}    & \multirow{2}{*}{\textbf{MobileNetv2}}    &
    \multirow{2}{*}{74.3}    & \multirow{2}{*}{73.4}    & \multirow{2}{*}{73.9}    & \multirow{2}{*}{73.8} \\
    {} & {} & {}    & {}    & {}    & {} \\
        \midrule[\heavyrulewidth]
    \multirow{2}{*}{\textbf{Tiny ImageNet}}    & \multirow{2}{*}{\textbf{ResNet18}}    &
    \multirow{2}{*}{65.2}    & \multirow{2}{*}{63.9}    & \multirow{2}{*}{64.6}    & \multirow{2}{*}{64.3} \\
    {} & {} & {}    & {}    & {}    & {} \\
        \midrule[\heavyrulewidth]
    \multirow{2}{*}{\textbf{ImageNet}}    & \multirow{2}{*}{\textbf{ResNet18}}    &
    \multirow{2}{*}{70.3}    & \multirow{2}{*}{69.7}    & \multirow{2}{*}{70.1}    & \multirow{2}{*}{70.1} \\
    {} & {} & {}    & {}    & {}    & {} \\ 
        \midrule[\heavyrulewidth]
    \multirow{2}{*}{\textbf{ImageNet}}    & \multirow{2}{*}{\textbf{ViT-B}}    &
    \multirow{2}{*}{83.8}    & \multirow{2}{*}{78.8}    & \multirow{2}{*}{82.3}    & \multirow{2}{*}{80.9} \\
    {} & {} & {}    & {}    & {}    & {} \\

    \bottomrule
\end{tabular}}
\end{table}
\begin{table}[tb]
\centering
\caption{{Mixed-Computation Performance on Language Models }}
\label{tab:language models}
\resizebox{\columnwidth}{!}{
\begin{tabular}{c c c c c c}
    \toprule
    \multirow{2}{*}{\textbf{Dataset}}   & \multirow{2}{*}{\textbf{Model}}    & \textbf{Full-Precision (FP32)} & \textbf{FixP4}         & \textbf{Posit4}         &  \textbf{Mixed-Computation} \\
    {}                                  & {}                                        & \textbf{(\%)}  & \textbf{(\%)}             & \textbf{(\%)}     & \textbf{(\%)} \\
    \midrule[\heavyrulewidth]
  
    \multirow{2}{*}{\textbf{MNLI}}    & \multirow{2}{*}{\textbf{BERT}}    &
    \multirow{2}{*}{84.1}    & \multirow{2}{*}{77.2}    & \multirow{2}{*}{83.7}    & \multirow{2}{*}{80.0} \\
    {} & {} & {}    & {}    & {}    & {} \\ 

        \midrule[\heavyrulewidth]
    \multirow{2}{*}{\textbf{SST-2}}    & \multirow{2}{*}{\textbf{BERT}}    &
    \multirow{2}{*}{90.8}    & \multirow{2}{*}{86.4}    & \multirow{2}{*}{89.7}    & \multirow{2}{*}{88.4} \\
    {} & {} & {}    & {}    & {}    & {} \\ 

        \midrule[\heavyrulewidth]
    \multirow{2}{*}{\textbf{WikiText-2}}    & \multirow{2}{*}{\textbf{GPT-2}}    &
    \multirow{2}{*}{3.40 (Loss)}    & \multirow{2}{*}{5.9 (Loss)}    & \multirow{2}{*}{3.45 (Loss)}    & \multirow{2}{*}{4.26 (Loss)} \\
    {} & {} & {}    & {}    & {}    & {} \\ 
    \bottomrule
\end{tabular}}
\vspace{-2em}
\end{table}

\begin{figure}
\vspace{-1em}
\centering
\includegraphics[scale=0.09]{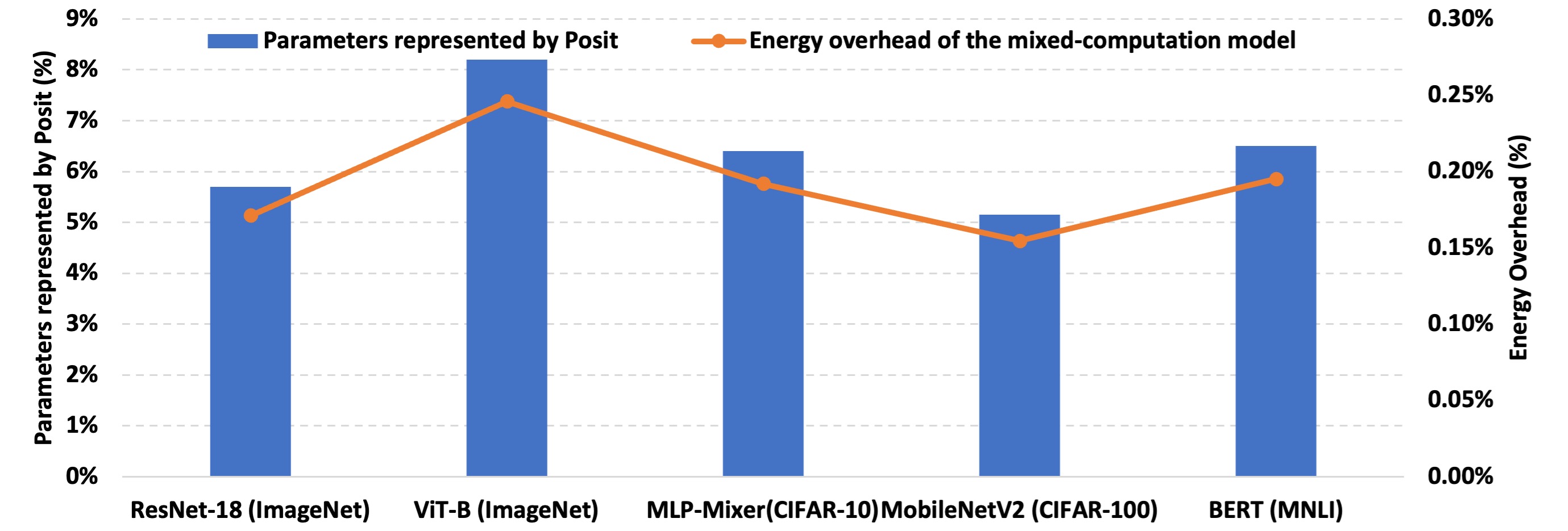}
\caption{The percentage of the parameters that are mapped to Posit in mixed-computation model and the imposed energy overhead in the mixed-computation apporach.} 
\label{result:energy}
\end{figure}

Finally, Figure \ref{result:energy} illustrates the percentage of parameters mapped to Posit representation for several models within our proposed framework. Additionally, this figure presents the energy overhead associated with Posit-based computation, assuming that the MAC operation is responsible for a maximum of 10\% of the total system energy consumption, as indicated in \cite{8686088}. Notably, based on the chosen threshold ($\eta=10\%$), the percentage of parameters utilizing Posit representation in all models remains below 10\%. In the worst-case scenario, the resulting energy overhead from mixed computation amounts to approximately 0.25\%.

